\pdfoutput=1

\documentclass[11pt]{article}

\usepackage{acl}

\usepackage{times}
\usepackage{latexsym}

\usepackage[T1]{fontenc}

\usepackage[utf8]{inputenc}

\usepackage{microtype}
\usepackage{amssymb}

\usepackage{inconsolata}

\usepackage{graphicx}
\usepackage{subcaption}
\usepackage{fontawesome}
\usepackage{booktabs}
\usepackage{comment}

\newcommand{\heart}{\ensuremath\heartsuit}
\newcommand{\diamondsmall}{\ensuremath\diamondsuit}

\title{SADAS: A Dialogue Assistant System Towards Remediating Norm Violations in Bilingual Socio-Cultural Conversations}

\author{Yuncheng Hua\textsuperscript{\rm \heart}, Zhuang Li\textsuperscript{\rm \heart}, Linhao Luo\textsuperscript{\rm \heart}, Kadek Ananta Satriadi\textsuperscript{\rm \heart}, Tao Feng\textsuperscript{\rm \heart}, \\ \textbf{Haolan Zhan}\textsuperscript{\rm \heart},  
\textbf{Lizhen Qu}\textsuperscript{\rm \heart}, \textbf{Suraj Sharma}\textsuperscript{\rm \diamondsmall},
\textbf{Ingrid Zukerman}\textsuperscript{\rm \heart}  \\
\textbf{Zhaleh Semnani-Azad}\textsuperscript{\rm \diamondsmall} \and
\textbf{Gholamreza Haffari}\textsuperscript{\rm \heart} \\
\textsuperscript{\rm \heart} Department of Data Science \& AI, Monash University, Australia\\
\textsuperscript{\rm \diamondsmall}  California State University, Northridge, CA \\
\{devin.hua, firstname.lastname\}@monash.edu, \\ \{suraj.sharma, zhaleh.semnaniazad\}@csun.edu\\ 
}

\begin{document}
\maketitle

\begin{abstract}
In today's globalized world, bridging the cultural divide is more critical than ever for forging meaningful connections.
The Socially-Aware Dialogue Assistant System (SADAS) is our answer to this global challenge, and it's designed to ensure that conversations between individuals from diverse cultural backgrounds unfold with respect and understanding.
Our system's novel architecture includes: (1) identifying the categories of norms present in the dialogue, (2) detecting potential norm violations, (3) evaluating the severity of these violations, (4) implementing targeted remedies to rectify the breaches, and (5) articulates the rationale behind these corrective actions. 
We employ a series of State-Of-The-Art (SOTA) techniques to build different modules, and conduct numerous experiments to select the most suitable backbone model for each of the modules.
We also design a human preference experiment to validate the overall performance of the system.
We will open-source our system (including source code, tools and applications), hoping to advance future research.
A demo video of our system can be found at:~\url{https://youtu.be/JqetWkfsejk}. We have released our code and software at:~\url{https://github.com/AnonymousEACLDemo/SADAS}.
%
\end{abstract}

\section{Introduction}

\begin{figure*}
    \centering
    \includegraphics[width=1\textwidth]{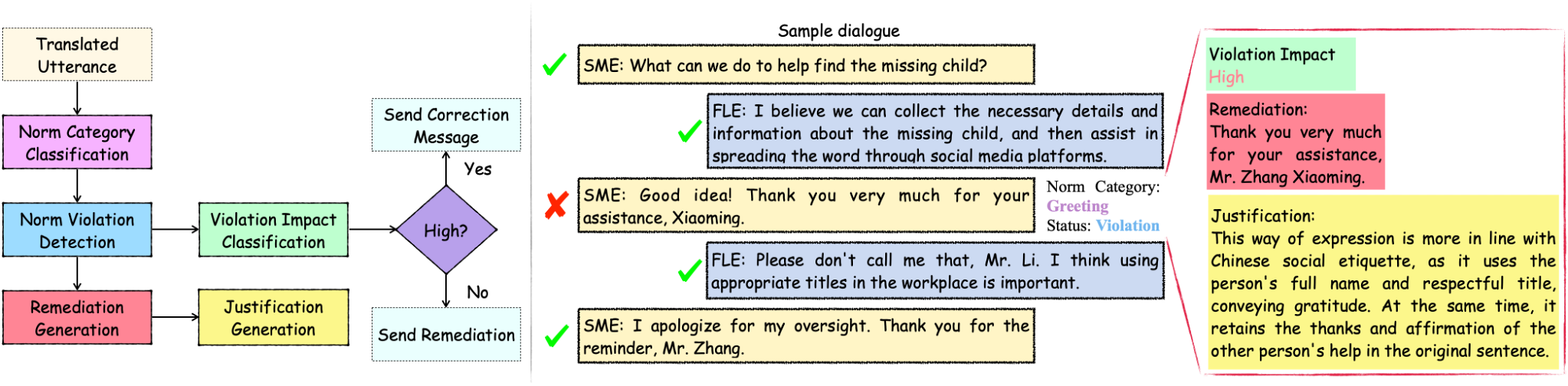}
    \caption{Main tasks of our framework (left): (1)~norm category classification, (2)~norm violation detection, (3)~violation impact classification, (4)~remediation generation, and (5)~justification generation. (right) A running example: In this conversational exchange between two interlocutors, an utterance from one party (referred to as SME) breaches a social norm. Our system, SADAS, identifies the breach and intervenes to rectify it by generating a remedial response and crafting a justification for its actions.}
    \label{fig:intro}
\end{figure*}

In an increasingly globalized world, individuals from diverse cultures are engaging, cooperating, and contending with each other more extensively than ever.
These distinct values and social norms from different cultures heavily influence people's objectives and behaviours in a variety of circumstances~\cite{morris2015normology}.
%
%
%
As a result, people with different cultural backgrounds may unintentionally fail to meet behavioural expectations due to insufficient understanding of each other's social norms during cross-cultural communication.
Based on the previous research on the punishment responses to norm violations~\cite{molho2020direct}, when social norm violations occur, it is common to see various forms of punishment, such as conflict, failed trade, and breakdown of negotiation etc.
%

In order to mitigate norm violations in \textit{cross-cultural} and \textit{bilingual} communication without involving cross-cultural human experts, it is desirable to build a dialogue assistant that can provide real-time assistance to negotiations that goes beyond machine translation~\cite{DBLP:conf/eamt/Pluymaekers22}.
The existing NLP systems struggle when it comes to understanding the social factors of language, restricting their use to culturally-aware negotiation systems~\cite{DBLP:conf/naacl/HovyY21}. 
Post-editing, a useful method of polishing the output of machine translation models, does not provide dialogue assistance in a cross-language and cross-cultural setting~\cite{DBLP:conf/emnlp/SartiBAT22}.  
Task-oriented and retrieval-based dialogue systems struggle with open-domain conversations~\cite{DBLP:journals/corr/abs-1907-12878,DBLP:conf/eacl/ChawlaSZLYG23}.
Deep generative dialogue models, state-of-the-art as they are, face challenges in accurately analyzing cross-cultural communication~\cite{guo2023aigc}. 

To provide cultural insight in the negotiation process, prior works build valuable resources related to social norms, such as NORMSAGE~\cite{DBLP:journals/corr/abs-2210-08604},  Moral Integrity Corpus ({MIC})~\cite{DBLP:conf/acl/ZiemsYWHY22} and PROSOCIALDIALOG corpus~\cite{DBLP:conf/emnlp/0002YJLKKCS22}. However, the models built on those datasets focus mainly on detecting norm related information. SOGO~\cite{DBLP:conf/iva/ZhaoRR18} could detect social norm violations, but failed to remedy them. Similarly, although the agents built in~\citep{DBLP:conf/emnlp/0002YJLKKCS22} can generate socially appropriate responses, they are not designed to detect norm violations. 

%
%
%
%
In this paper, we propose Socially-Aware Dialogue Assistant System (SADAS), which is the \textit{first} system for detecting, assessing and remedying norm violations occurring in bipartite and bilingual conversations.
As illustrated in Figure~\ref{fig:intro}, our system accomplishes the following tasks sequentially: (1) recognizing the categories of the norms that occur in a dialogue, (2) detecting possible norm violations, (3) judging whether the detected norm violations have high or low impact, (4) remedying the norm violations and (5) justifying the remediation measures.

\begin{figure*}
    \centering
    \includegraphics[width=0.80\textwidth]{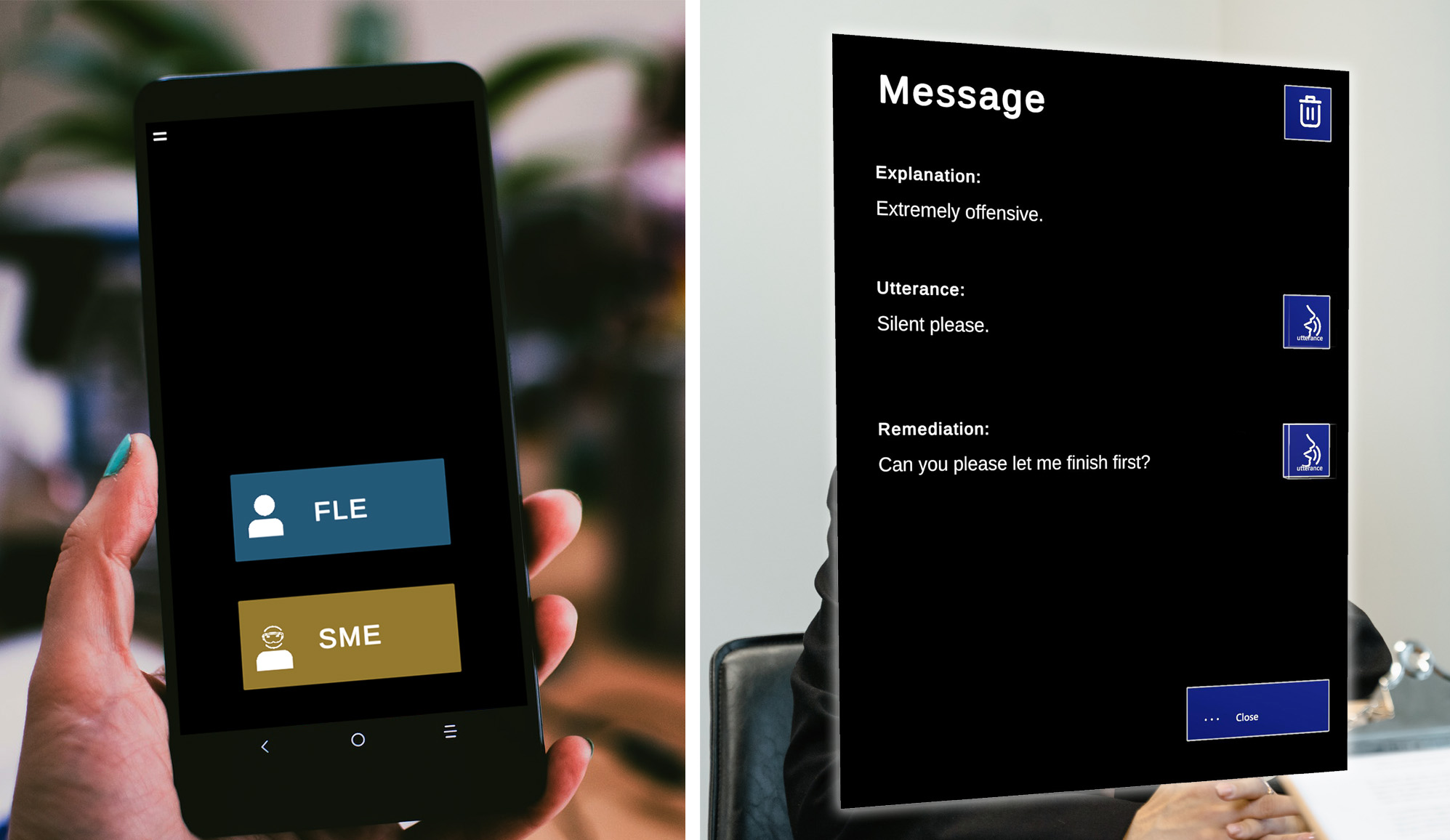}
    \caption{The interaction interface on the mobile application (left) and a sample correction message popping up in HoloLens 2's mixed-reality application (right).}
    \label{fig:mobile_hl2}
\end{figure*}

In a bilingual dialogue, an interlocutor wearing a Hololens 2 (sender) talks to another interlocutor (receiver) from the target culture. Neither interlocutor understands the other's language and culture. A machine translation (MT) module in our system translates a speech into the target language after applying Automatic Speech Recognition (ASR). If the norm violation detector identifies a norm violation, the norm category detector associates it with a norm category, which characterizes the intention of the sender and provides valuable information for the remediation generator. In the sequel, the generator suggests a remediation message that conveys the translation in a culturally more appropriate way and explains the message with the corresponding social norm knowledge. If the violation is classified as low impact, our system sends the remediation message directly to the text-to-speech module, otherwise, it sends the remediation message, the original utterance, and the justification to the Hololens screen and let the sender choose if he/she would like to play the remediation message or the translation. Fig. \ref{fig:mobile_hl2} shows an example Hololens message.
%
%
We develop modules to complete each of the tasks mentioned above and integrate them into a pipeline. The modules are trained and evaluated on SocialDial~\cite{zhan2023socialdial} and a small corpus constructed by ourselves. Please refer to Section~\ref{sec:experiment} to read more details about the training corpus.
%
%

%
To sum up, SADAS has the following highlights:
\begin{itemize}
  \item SADAS is the first real-time system to facilitate cross-cultural bilingual conversations between two interlocutors with different cultural backgrounds by detecting and remedying social norm violations. 
 \item We conduct both automatic and human evaluations for the whole system as well as the key modules. For each task mentioned above, we select the best possible module and justify our design choices based on empirical evidence.
\end{itemize}



\section{System Description}
\label{sec:system_description}
SADSA is applicable in the following scenario: both interlocutors in the conversation use different languages and cannot understand each other's language.
During the conversation, SADSA first translates the sender's speech into the language used by the receiver.
If no norm violation is detected, SADSA directly sends the translated sentence to the receiver.
However, if a norm violation is detected, SADSA generates a remediation message and interacts with the sender to decide whether to send the translated sentence or the remediation message to the receiver.

\subsection{System Architecture}
As mentioned above, we designed a series of pipeline tasks to achieve the goal of norm violation recognition and remediation.
We designed modules for each of the tasks respectively, including norm category classifier, norm violation classifier, norm violation impact classifier, remediation generator, and justification generator.
We collectively refer to these modules as the \textbf{\textit{core}} modules.
In addition to the \textbf{\textit{core}} modules, we also designed other \textbf{\textit{function}} modules including input/output (I/O) modules, message middleware, Automatic Speech Recognition (ASR), and Machine Translation (MT).
The \textbf{\textit{core}} modules and \textbf{\textit{function}} modules interact and collaborate through the main engine, thereby constructing an organic system.
We establish the main engine using a pipeline manner based on a finite-state machine (FSM)\footnote{\url{https://en.wikipedia.org/wiki/Finite-state_machine}}.

\subsection{Applications}
Our I/O modules consist of two main applications: (1) a mobile application (Android), (2) a mixed-reality (MR) application run on HoloLens 2 headset\footnote{\url{https://www.microsoft.com/en-au/hololens}}. We built these applications using Unity Engine version 2019.4.14f\footnote{\url{https://unity.com/}}. The mobile application has two buttons labelled ``Subject Matter Expert (SME)'' and ``Foreign Language Expert (FLE)'', representing the entity of the two interlocutors. 

%
%

\subsection{Process}
\label{subsec:process}
Before the two interlocutors engage in a dialogue using the SADAS system, they choose their respective identities on the mobile application (Figure~\ref{fig:mobile_hl2}).
%
%
When speaking, users press and hold their corresponding buttons and release them when finished speaking.
The above press-to-talk (PTT) mechanism is a simple but effective design that averts the speaker diarisation problem existing in the process of partitioning an audio stream with human speech into separate segments based on the speaker's identities~\cite{DBLP:journals/csl/ParkKDHWN22}.
The recorded speech, along with its identity token, is transmitted from the mobile end to the main engine through the message middleware.
We employ a two-way interactive WebSocket API\footnote{\url{https://pypi.org/project/websocket-client/}} to build the message middleware.
As the main engine receives the speech, it calls an ASR model, being built on Whisper~\cite{DBLP:journals/corr/abs-2212-04356}, to transcribe the speech into a textual sentence, and calls a MT model (by using Azure AI Translator\footnote{\url{https://learn.microsoft.com/en-us/azure/ai-services/translator/}}) to translate the sentence.
The translated sentence and its corresponding identity token are added to the dialogue history, and are further managed by the main engine.

Then, the main engine calls the \textbf{\textit{core}} modules to accomplish the tasks in a pipeline manner.
The norm category/violation classifiers first identify the possible norm violations in the incoming textual sentences.
If a norm violation is detected, the norm violation impact classifier is used to measure the severity of the norm violation.
Meanwhile, the remediation generators rewrites the sentence that triggers norm violations and rectifies the violations without altering the translated sentence's meaning.
The justification generator then composes a justification message to explain the necessity of the rewriting and the specific social norms and cultural etiquette on which the rewriting is based.

Specifically, we categorize the norm violations into \textbf{\textit{low}}-impact and \textbf{\textit{high}}-impact violations. 
We define \textbf{\textit{high}}-impact norm violations as the violations that more likely lead to the receiver's negative emotion, the damage to the two interlocutors' relationship, or even the breakdown of negotiation.
As shown in figure~\ref{fig:intro}, provided a \textbf{\textit{low}}-impact violation, our system will automatically send the remediation rather than the translated sentence to the receiver.
On the contrary, once a \textbf{\textit{high}}-impact violation is detected, the system will send the correction message to the sender's Hololens 2 interface, and wait for the sender's choice.
As depicted in Figure~\ref{fig:mobile_hl2}, by pressing the button beside each sentence, the sender can choose from the translated sentence and the remediation to better convey his/her original meaning in a more norm-adherence way to the receiver.
After the sender presses a button, the sender's choice will be sent to the main engine, and the engine thus sends the translated sentence or the remediation message to the receiver conditioning on the sender's choice.    

\subsection{Norm Category Classification and Violation Classification}

The first \textbf{\textit{core}} module of SADAS is to predict norm categories and detect norm violations in dialogue utterances.
%
%
As proposed by ~\citet{zhan2023socialdial}, the categorization of norms is based on individuals' intents, including seven norm categories.
%
Utterances that do not engage with these specified norms are classified as \textit{other}.
Given an utterance associated with a norm category, the norm violation detection task then assesses whether the utterance \textit{adheres to} or \textit{violates} the associated norm.
Norm category prediction is treated as a multi-way classification task, while norm violation detection is viewed as a binary classification task.
We employed ChatGPT~\cite{openai22} and a bilingual ChatGPT-like model ChatYuan~\cite{clueai2023chatyuan} as the backbone models.
We provided ChatGPT with multiple in-context learning samples (dialogue sentences with annotated norm categories and violation labels) in the prompt, followed by the current sentence in the dialogue, and asked ChatGPT to conduct the classification tasks.
Also, we fine-tuned ChatYuan using Socialdial to develop two classification models.
As depicted in~\ref{subsec:norm_category_violation}, we chose the optimal setting as our classifier. 

\subsection{Norm Violation Impact Classification}
The norm violation impact module aims to evaluate the degree of norm violation as \textbf{\textit{high}} or \textbf{\textit{low}}. The evaluation of norm impact is based on the severity of offending the interlocutor. The impact of a violation is considered high if it would lead to serious consequences, such as a relationship breakdown. As shown in Figure~\ref{fig:intro}, the estimation of violation impact is formulated as a binary classification task.
%

\subsection{Generation of Remediation and Justification}
The process of creating remediation, along with the justification for such remediation, is first conducted by GPT-3.5~\cite{openai22}.
We provide GPT-3.5 with ten human-written pairs of utterances that require remediation, along with their appropriate remedial counterparts.
These examples serve as part of the task instruction. Upon sending an instruction through the GPT-3.5 API~\cite{openai23}, the model generates a suitable remediation for any offensive utterance and explains the proposed remediation.
Following this, we employ regular expressions to extract the resulting remediation and its associated explanation accurately.

Furthermore, considering the potential instability in calling the GPT-3.5 API, such as high network latency or occasional errors during peak API requests, we implemented a local ChatYuan~\cite{clueai2023chatyuan} as a backup in SADAS.
Once there is no response within a certain period of time, or an error occurs while calling the GPT-3.5 API, we will activate the local ChatYuan to generate remediation and justification.
This ensures that the system can still function smoothly and provide timely responses even in situations where API calls may not be reliable.
We use SocialDial dataset to finetune ChatYuan by regarding the task instruction followed by a multi-turn dialogue as the input and the ground-truth remediation/justification as the target output.
To minimize the training cost, we adopt adapter-based tuning as a parameter-efficient alternative~\cite{he2021effectiveness}. 
When testing, upon sending an instruction to the tuned LLM, ChatYuan generates a suitable remediation for any offensive utterance and explains the proposed remediation.

\section{Experimental Details}
\label{sec:experiment}
We use the SocialDial dataset for training SADSA's \textbf{\textit{core}} modules.
The training set of Socialdial consists of two different distributions of data: Human-Authored (manually written by professional annotators) and Synthetic Data (generated by GPT-3.5 based on instruction-based prompts).
The test set of Socialdial is entirely manually composed.

The SocialDial corpus is chosen because it is currently the only socially-aware dialogue dataset available in existing work.
It comprises various norm categories and topics to enhance the diversity of dialogues and reduce potential bias issues.
The socially-aware factors, including social norm category, norm remediation, and norm justification, are explicitly annotated in the dataset.
Note that while SocialDial is a Chinese dataset, the social norm, culture, and phenomena contained in its dialogues are language-agnostic.
Such the social factors are not specific to Chinese culture but can be applied universally across different languages and cultures.
%
%

\subsection{Task 1 and 2: Norm Category Classification and Violation Classification}
\label{subsec:norm_category_violation}
\paragraph{Experimental Settings.} 
%
We implemented ChatYuan\footnote{\url{https://github.com/clue-ai/ChatYuan}}, fine-tuned it using the SocialDial dataset, and implemented ChatGPT by composing prompts and calling the GPT-3.5 API.
We also proposed an \textit{ensemble} setting, where we applied a stacking method~\cite{dvzeroski2004combining} to combine the predictions of both the ChatYuan and ChatGPT models by using a linear classifier.
Specifically, for an query, we first sent it to both ChatYuan and ChatGPT models to get thier predictions. We converted the discrete output of ChatGPT into an 8-dimensional one-hot vector (with 7 norm categories and one \textit{other} category). At the same time, ChatYuan also generated an 8-dimensional probability distribution over eight categories for this query.
We employed a linear classifier, taking the concatenation of the two vectors as features, to learn how to improve the classification performance for both the norm category and violation tasks.
We trained ChatYuan on both synthetic and human-authored training sets, and tested both ChatYuan and ChatGPT using the human-authored test set.  
Our evaluation metrics for these experiments included precision, recall, and the F1 score.

\noindent \textbf{Discussion.} 
We present the experimental results of norm category prediction and norm violation detection in Table~\ref{tab:norm_category_detection_results} and Table~\ref{tab:norm_violation_detection_results}
respectively.
Our findings indicate that ChatGPT could get superior results compared to other settings on the norm category prediction task.
Also, the ensemble mechanism was able to achieve the best F1-Micro score on the norm violation detection task.
Therefore, we chose ChatGPT and ensemble as SADAS's norm category and violation classifiers, respectively.

\begin{table}[!t]
\centering
\scriptsize
\begin{tabular}{llll}
\hline
\textbf{Model} & \textbf{P} & \textbf{R} & \textbf{F1-Micro} \\ \hline
ChatYuan       & 0.473      & 0.528      & 0.498             \\
ChatGPT        & \textbf{0.753}      & \textbf{0.670}      & \textbf{0.709}             \\
Ensemble       & 0.534      & 0.545      & 0.539             \\ \hline
\end{tabular}
\caption{Experiment results of norm category prediction.}
\label{tab:norm_category_detection_results}
\end{table}

\begin{table}[!t]
\centering
\scriptsize
\begin{tabular}{llll}
\hline
\textbf{Model} & \textbf{P} & \textbf{R} & \textbf{F1-Micro} \\ \hline
ChatYuan       & 0.590      & \textbf{0.684}      & 0.633             \\
ChatGPT        & 0.36       & 0.473      & 0.411             \\
Ensemble       & \textbf{0.812}      & 0.681      & \textbf{0.741}             \\ \hline
\end{tabular}
\caption{Experiment results of norm violation detection.}
\label{tab:norm_violation_detection_results}
\end{table}

\subsection{Task 3: Norm Violation Impact Classification}

\begin{table}[!t]
\centering
\scriptsize
\begin{tabular}{lccc}
\hline
  \textbf{Models}    & 
 \textbf{P}   & \textbf{R}   & \textbf{F1}   \\ \hline
 BERT-zh  & 72.48 & 69.43 & 70.92              \\
 RoBERTa-zh  & \textbf{80.16} & \textbf{76.59} &  \textbf{78.33}        \\ \hline
\end{tabular}%
\caption{Experiment results of impact estimation of violation models trained on three different training settings.}
\label{tab:impact_violation}
\end{table}


\noindent \textbf{Experimental Settings.}
We have formulated the estimation of violation impact as a binary classification task. 
We utilized BERT-zh~\cite{devlin-etal-2019-bert} and RoBERTa-zh~\cite{liu2020roberta} as two baseline models. Precision, recall, and the F1-score were used as evaluation metrics. Given that the previous dialogue context could influence impact estimation, we concatenated the current violation utterance with the two preceding utterances. This composite input was then fed into the classifier.

\noindent \textbf{Discussion.} 
We present the results of violation impact estimation in Table~\ref{tab:impact_violation}.
RoBERTa-zh outperforms BERT-zh to a large extent. Therefore, we finally choose RoBERTa-zh as the impact estimation model in our system.

\subsection{Task 4 and 5: Generation of Remediation and Justification}
\paragraph{Experimental Settings.} 
To find the optimal adapter for the ChatYuan model, we conduct experiments to compare the performance of different settings.
We implemented ChatYuan and paired them with two adapters: \textit{Pfeiffer}~\cite{pfeiffer2020adapterfusion} and \textit{Prefix tuning}~\cite{li2021prefix}. 
We used both the Socialdial's human-authored and synthetic training data to finetune the models, and evaluated them on the human-authored test data.
We used automatic evaluation metrics, including BLEU~\cite{papineni2002bleu}, ROUGE-L (using F1)~\cite{lin2004rouge}, MAUVE~\cite{pillutla2021mauve}, and BERT-Score (using F1)~\cite{zhang2019bertscore}, to measure the difference between models' outputs and manually crafted reference ones.
We qualitatively compared the outputs of different settings from three perspectives: Effectiveness (\textit{Effect.}), Relevance (\textit{Rel.}) and Informative (\textit{Info.}).
%
Human annotators scored the outputs with 1 indicating the lowest preference and 3 representing the highest. 

\begin{table}[!t]
    \centering
    \tiny
    \begin{tabular}{lccccc}
    \toprule
      \multicolumn{6}{c}{\textbf{Remediation Generation}} \\ 
       Model  &  BLEU. & R-L & MAUVE & BScore & Avg. Len\\ \hline
        ChatYuan + \textit{Pfeiffer}  & \textbf{0.244} & \textbf{0.359} & \textbf{0.384} & \textbf{0.713} & 28.78    \\
          ChatYuan + \textit{Prefix tuning}  & 0.161 & 0.311 & 0.280 & 0.699 & 17.93    \\ \hline\midrule
         \multicolumn{6}{c}{\textbf{Justification Reason Generation}} \\ 
      Model  &  BLEU. & R-L & MAUVE & BScore & Avg. Len\\ \hline
       ChatYuan + \textit{Pfeiffer}  &\textbf{0.106} & 0.150 & 0.014 & 0.603 & 66.46     \\
          ChatYuan + \textit{Prefix tuning}  & 0.103 & \textbf{0.154} & 0.014 & \textbf{0.611} & 58.10   \\
    \bottomrule
    \end{tabular}
    \caption{Automatic evaluation on the remediation generation and justification reason generation tasks.}
    \label{tab:auto_reme}
\vspace{-3mm}
\end{table}

\noindent \textbf{Discussion.} 
From Table~\ref{tab:auto_reme}, it can be observed that ChatYuan+\textit{Pfeiffer} performed better for all metrics on the remediation generation task.
Meanwhile, in terms of justification generation, ChatYuan+\textit{Pfeiffer} and ChatYuan+\textit{Prefix tuning} achieved similar performance.
The results of human preference reported in Table~\ref{tab:human_reme} shows that ChatYuan+\textit{Pfeiffer} performed better in both the tasks.
These findings verify that ChatYuan+\textit{Pfeiffer} can generate remediation that better align with the annotators' understanding of social norms.
We thus chose ChatYuan+\textit{Pfeiffer} as our system's backup remediation and justification generators.    

\begin{table}[!t]
    \centering
    \tiny
    \begin{tabular}{lcccc}
    \toprule
      \multicolumn{5}{c}{\textbf{Remediation Generation}} \\ 
       Model  &  Effect. & Rel. & Info. & \textit{kappa} \\ \hline
        ChatYuan + \textit{Pfeiffer}  &  \textbf{2.29} & 	\textbf{2.71} & 	\textbf{2.79} &  0.49   \\
          ChatYuan + \textit{Prefix tuning}  & 1.94 & 2.35 & 2.16 &   0.56  \\ \hline\midrule
         \multicolumn{5}{c}{\textbf{Justification Reason Generation}} \\ 
       Model  &  Effect. & Rel. & Info. & \textit{kappa} \\ \hline
        ChatYuan + \textit{Pfeiffer}  &  \textbf{2.46} & \textbf{2.59} & \textbf{2.72} & 0.55  \\
          ChatYuan + \textit{Prefix tuning}  & 2.14 & 2.48 & 2.25   & 0.57 \\
    \bottomrule
    \end{tabular}
    \caption{Human evaluation on the remediation generation and justification reason generation task respectively.}
    \label{tab:human_reme}
\end{table}

\subsection{Task 6: User Study}

As described in Section~\ref{subsec:process}, in high-impact norm violation scenarios, we provide two options for the sender to choose from, i.e., translated sentence or remediation.
%
To evaluate the quality of remediation measures generated by the whole system, we ask a group of experiment participants to use our system to conduct 40 bilingual dialogues related to Chinese social norms.
In each dialogue, one interlocutor wears the Hololens and speaks English (we view this interlocutor as SME), while the other interlocutor speaks Mandarin (FLE).
Herein, both interlocutors are provided with a background of the dialogue beforehand and are instructed to play the corresponding roles as naturally as possible. Each dialogue consists of approximately 30 turns.

If our system detects an utterance violating a norm with high impact, SME chooses the remediation measure only if the remediation measure is culturally more appropriate than the translated sentence and the remaining part preserves the corresponding meaning of the original utterance.
Therefore, the ratio of choosing remediation measures is considered a key metric for assessing the quality of generated remediation measures.

Among the 40 dialogues, our system detects 25 low-impact and 117 high-impact violations.
Hololens users selected remediations for 56 of the 117 high-impact violations, demonstrating that our system is capable of mitigating culturally inappropriate behaviors in close to 48\% of the instances where such high-impact violations were detected.
Although ChatGPT is considered one of the strongest LLMs for this task, there is still significant room for improvement. 

Furthermore, we computed the average system response time (the time spent on translating and remedying the sentence).
Due to the unpredictable response time caused by user-initiated selection in high-impact remediation scenario and the unstable GPT-3.5 web service calls, we use local models (RoBERTa-zh and ChatYuan) to do the inference under the two specific scenarios: those with no remediation and those with low-impact remediation.
According to the data recorded, the mean response time is 1.5 seconds in the absence of remediation, whereas it extends to 6.7 seconds when low-impact norm remediations are required.
Reducing system latency is one of our future priorities.


\section{Conclusion}
SADAS is a socially-aware dialogue system that assists the interlocutors with different cultural and linguistic backgrounds in having conversations without barriers, misunderstandings, or even conflicts.
It is built with well-designed core modules (to detect, evaluate, and remedy norm violations), function modules (including I/O applications, message middleware, ASR, and MT), and main engine.
SADAS incorporates SOTA NLP technologies to achieve its functionality and provides users with a straightforward and user-friendly interfaces to operate it.
We encourage researchers, the NLP community, and linguists to use, extend, and contribute to this socially-aware dialogue system.

\section*{Ethics Statement}
We outline several ethical considerations to address and mitigate potential risks.

\paragraph{Temporal bias.}
In this study, our approach to recognizing and remediating social norms involves the utilization of pre-trained language models (LLMs), such as ChatGPT, ChatYuan, and ChatGLM.
These models serve as powerful tools to gain actionable insights regarding the regulations and evaluations of acceptable conduct from human conversational interactions.
It is important to acknowledge that social, socio-cultural, and moral norms can exhibit contextual variations over time.
Therefore, our discernment of norms is pertinent to the specific time frame corresponding to the conversational context in which a norm is identified.

\paragraph{Cultural bias.}
It is crucial to note that LLM models acquire their implicit knowledge from vast datasets and incorporate mechanisms to mitigate bias.
However, it is imperative to recognize that all computational models inherently carry a risk of potential bias, particularly in relation to diverse languages and cultures.
We strongly encourage researchers and practitioners to exercise caution and implement safeguards in their endeavors.

\paragraph{Norm Content Clarification.}
It is imperative to recognize that the automated generation of norm remediations and justifications may be construed as normative and authoritative in nature.
Nonetheless, it is of paramount importance to elucidate that we do not regard the identified norm-relevant contents as universally applicable or globally binding standards.
The purpose of these norms is not to establish a comprehensive and all-encompassing ethical framework; instead, they serve as discrete intuitions and references.

\paragraph{Misuse of System.}
The primary objective of this system is to identify and remediate offensive language, especially when an interlocutor may not be familiar with one another's cultural backgrounds in a conversation. We strongly emphasize that this system must not be employed to create or spread content that could discriminate against or harm any cultural group.

\paragraph{Risks in User Study.}
We highly value the welfare and fair treatment of the participants in our user study. They were justly compensated at a rate of 15 USD per hour throughout the research process. Recognizing the potential sensitivity of the material in the study cases, we implemented protocols that provide for necessary breaks, enabling any participant who felt discomfort to take time away as needed.


\bibliography{anthology,custom}

\end{document}